\documentclass[journal]{IEEEtran}
\ifCLASSINFOpdf
  \usepackage[pdftex]{graphicx}
\else
\fi
\usepackage[cmex10]{amsmath}
\usepackage{amsfonts}
\usepackage{cite}
\usepackage{hyperref}
\usepackage{multirow}
\usepackage[table,xcdraw]{xcolor}

\usepackage{algorithmic}
\usepackage{orcidlink}
\usepackage{array} 
\usepackage[T1]{fontenc} 
\usepackage{lmodern}

\begin{document}
%
\title{Bayesian Transformer for Pan-Arctic Sea Ice Concentration Mapping and Uncertainty Estimation using Sentinel-1, RCM, and AMSR2 Data}
%
%
%

\author{Mabel Heffring \orcidlink{0009-0007-5069-7431},  Lincoln Linlin Xu \orcidlink{0000-0002-3488-5199},~\IEEEmembership{Member,~IEEE} 
\thanks{This work was supported by the Natural
Sciences and Engineering Research Council of Canada (NSERC) under Grant RGPIN-2019-06744.}
\thanks{Mabel Heffring and Lincoln Linlin Xu are with the Department of Geomatics Engineering, University of Calgary, Canada (email: (mabel.heffring1, lincoln.xu)@ucalgary.ca) (Corresponding author: Lincoln Linlin Xu)}.

}

%
%


\markboth{Journal of \LaTeX\ Class Files,~Vol.~13, No.~9, September~2014}%
{Shell \MakeLowercase{\textit{et al.}}: Bayesian Transformer for Pan-Arctic Sea Ice Concentration Mapping and Uncertainty Quantification using Sentinel-1, RCM, and AMSR2 Data }
%



\maketitle

\begin{abstract}

Although high-resolution mapping of Pan-Arctic sea ice with reliable corresponding uncertainty is essential for operational sea ice concentration (SIC) charting, it is a difficult task due to some key challenges, e.g., the subtle nature of ice signature features, model uncertainty, and data heterogeneity. This letter presents a novel Bayesian Transformer approach for Pan-Arctic SIC mapping and uncertainty quantification using Sentinel-1, RADARSAT Constellation Mission (RCM), and Advanced Microwave Scanning Radiometer 2 (AMSR2) data. First, to improve feature extraction, we design a novel high-resolution Transformer model with both global and local modules that can better discern the subtle differences in sea ice patterns. Second, to improve uncertainty quantification, we design a Bayesian extension of the proposed Transformer model, treating its parameters as random variables to more effectively capture uncertainties. Third, to address data heterogeneity, we fuse three different data types (Sentinel-1, RCM, and AMSR2) at decision-level to improve both SIC mapping and uncertainty quantification. The proposed approach is tested on Pan-Arctic datasets from September 2021, and the results demonstrate that the proposed model can achieve both high-resolution SIC maps and robust uncertainty maps compared to other uncertainty quantification approaches.

\end{abstract}


%
\IEEEpeerreviewmaketitle

\section{Introduction}

Accurate mapping of Pan-Arctic sea ice concentration (SIC) and evaluation of the SIC map uncertainties are essential for climate change studies, Arctic sea route navigation, and climate adaptation of the people and animals that call the Arctic home. Nevertheless, achieving high-resolution SIC maps with robust uncertainty quantification is a difficult task due to several key factors, i.e., the subtle nature of ice signature features, model ambiguity, and data heterogeneity. 

First, subtle differences in SIC classes can be difficult to detect in sea ice satellite images, including Synthetic Aperture Radar (SAR) from RADARSAT Constellation Mission (RCM) and Sentinel-1, and Passive Microwave (PM) radiometry from the Advanced Microwave Scanning Radiometer 2 (AMSR2), due to noise and the complex and ever-changing marine environment \cite{ wang2016sea, malmgren2021convolutional}. Therefore, the development of advanced deep learning approach to extract discriminative features from SAR and AMSR2 images is a critical task. Over the past decade, numerous deep learning (DL) approaches have been proposed. Convolutional Neural Networks (CNN) \cite{wang2016sea, wulf2024panarctic, malmgren2021convolutional} are commonly applied to sea ice concentration mapping due to their efficiency, but their strong inductive bias and focus on local features make it challenging to capture global context in SAR and PM sea ice imagery. In contrast, Transformers are more flexible and globally aware \cite{He2025physically}, making them a viable choice for learning the subtle nature of ice signature features on a Pan-Arctic scale. The development of high-resolution Transformer architecture is essential for improving sea ice feature extraction.

Second, model uncertainties cause significant ambiguities in the derived SIC maps. Variational inference methods such as Monte Carlo (MC) simulation, ensemble generation, and Bayes by Backpropagation (BBB) \cite{ABDAR2021243} are commonly used for DL uncertainty quantification in earth observation. Although MC simulation and ensemble generation can provide meaningful uncertainty estimates, these values are heuristic approximations that may not represent the true posterior distribution. Bayesian DL architectures that incorporate BBB explicitly estimate the model parameters as random variables \cite{aires2004neural}, resulting in more reliable and better calibrated uncertainties. Incorporating Bayesian uncertainty estimation into DL may also improve model performance and reduce misclassifications \cite{asadi2020evaluation}. 

Third, Pan-Arctic SIC mapping is complicated by Sentinel-1, RCM, and AMSR2 data heterogeneity. Various data fusion methods have been developed to combine the strengths of SAR and passive microwave systems for sea ice mapping, e.g., \cite{wang2016improved, wulf2024panarctic}. These methods are feature-level fusion approaches. The decision-level fusion, where fusion is performed after classification for each respective system, has not been fully explored to improve high resolution pan-Arctic sea ice mapping.

This letter presents a novel Bayesian Transformer approach for Pan-Arctic SIC mapping and uncertainty quantification using Sentinel-1, RCM, and AMSR2 data, with the following contributions.

\begin{itemize}
    \item First, to improve feature extraction, we design a novel high-resolution Transformer model with both global and local modules that can better discern the subtle differences in sea ice patterns.
    \item Second, to improve uncertainty quantification, we design a Bayesian extension of the proposed Transformer model, treating its parameters as random variables to more effectively capture uncertainties. 
    \item Third, to address data heterogeneity, we fuse three different data types (Sentinel-1, RCM, and AMSR2) at decision-level to improve both Pan-Arctic SIC mapping and uncertainty quantification.
\end{itemize}

 This letter proceeds as follows. Section II outlines the details of the Bayesian Transformer architecture and implementation. Section III presents the experimental design and results. Section IV concludes this study.

\section{Methodology}
\label{methodology}

\subsection{Dataset Overview}
Sentinel-1 is collected in Extra Wide (EW) mode with HH and HV polarization and 20 m x 40 m spatial resolution. RCM is collected in low resolution (100 m resolution), medium resolution (50 m), and low noise (100 m) modes with HH and HV polarization. AMSR2 is collected for 18.7 (10 km resolution), 23.8 (10 km), 36.5 (10 km), and 89.0 GHz (5 km) frequencies with H and V polarization. 18.7, 23.8, and 36.5 GHz are used to derive SIC using the NT algorithm \cite{NTdetails}, while 89.0 GHz is input for training. The model is weakly-supervised using the low resolution NT SIC and NIC ice charts. The data are processed into 256 x 256 image chips for 10 days in September 2021. Seven days are used for training (September 1st, 7th, 10th, 14th, 21st, 24th, and 30th), and three days are used for test (September 4th, 18th, and 27th). 

\subsection{High-Resolution Transformer Architecture}
\begin{figure*}[!htbp]
    \centering
    \includegraphics[scale = 0.32]{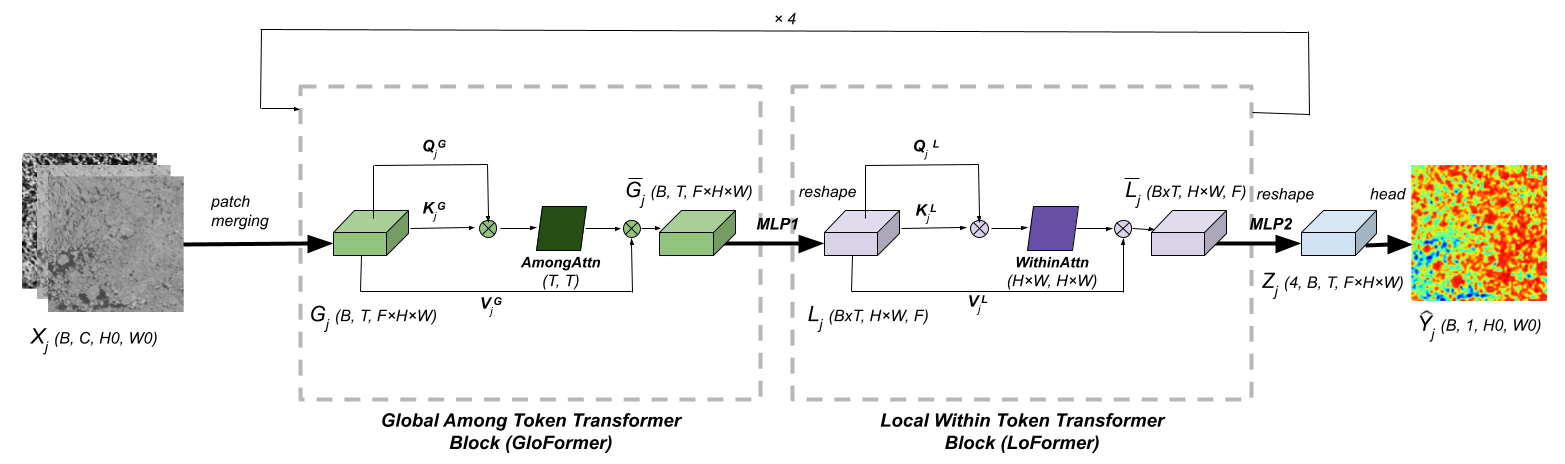}
    \vspace{-0.3cm}
    \caption{Proposed high-resolution Transformer architecture, $f^{\omega}(.)$, with an among token Transformer block for modeling global context (GloFormer) and a within token Transformer block for modeling local detail (LoFormer).}
    \label{fig:Transformer}
\end{figure*}

The high-resolution Bayesian Transformer is denoted as a regression function, $y = f^{\omega}(x)$, which derives SIC, $y$, from input data, $x$, by optimizing model parameters, $\omega$, through training. In Figure \ref{fig:Transformer}, the input data for the $j^{th}$ batch is $X_j \in \mathbb{R}^{B, C, H0, W0}$ where $B$ is the batch size, $C$ is the channel dimension, and $H0, W0$ are the height and width of the input chips, respectively. Patch merging is used to generate patch tokens ($G_j \in \mathbb{R}^{B, T, F \times H \times W}$), 
where $T$ is the number of tokens, $F$ is the hidden dimension, and $H,W$ are the height and width of each token, respectively. 

The among token Transformer block (GloFormer) models the global context using a multi-head self-attention mechanism across all tokens. For the first head, the output is defined as:
\begin{equation}
\bar{G}_j^1 = \text{softmax}\Bigg(\frac{Q_j^G (K_j^G)^\top}{\sqrt{d_k^G}}\Bigg) V_j^G = (AmongAttn)V_j^G
\label{global_attn_eq}
\end{equation}
where $Q_j^G = G_j \mathcal{W}^{Q_j^G}, K_j^G = G_j \mathcal{W}^{K_j^G}, V_j^G = G_j \mathcal{W}^{V_j^G} \in \mathbb{R}^{B\times T\times d_k^G}$ are queries, keys, and values of the tokens, respectively, $d_k^G = (F\times H\times W) / h$ is the GloFormer key dimension, $h$ is the number of heads, and $AmongAttn$ are the attention scores. $\mathcal{W}^{Q_j^G}$, $\mathcal{W}^{K_j^G}$, and $\mathcal{W}^{V_j^G}$ are the projection weights of $Q_j^G$, $K_j^G$, and $V_j^G$, respectively. Concatenated and projected heads, $[\bar{G}_j^1, \bar{G}_j^2,...,\bar{G}_j^h]$, yield $\bar{G}_j \in \mathbb{R}^{B, T, F \times H \times W}$.

The within token Transformer block (LoFormer) is designed to learn the local features using a multi-head self-attention mechanism across the patches inside each token. The output $\bar{G}_j$ is reshaped to $L_j \in \mathbb{R}^{B \times T, H \times W, F}$, and the LoFormer attention matrix for the first head is defined as:
\begin{equation}
\bar{L}_j^1 = \text{softmax}\Bigg(\frac{Q_j^L (K_j^L)^\top}{\sqrt{d_k^L}}\Bigg) V_j^L = (WithinAttn)V_j^L
\label{local_attn_eq}
\end{equation}
where $Q_j^L = L_j \mathcal{W}^{Q_j^L}, K_j^L = L_j \mathcal{W}^{K_j^L}, V_j^L = L_j \mathcal{W}^{V_j^L} \in \mathbb{R}^{B \times T \times d_k^L}$ are queries, keys, and values of the inner patches, respectively, $d_k^L = F / h$ is the LoFormer key dimension, $h$ is the number of heads, and $WithinAttn$ are the attention scores. $\mathcal{W}^{Q_j^L}$, $\mathcal{W}^{K_j^L}$, and $\mathcal{W}^{V_j^L}$ are the projection weights of $Q_j^L$, $K_j^L$, and $V_j^L$, respectively. Concatenated and projected heads, $[\bar{L}_j^1, \bar{L}_j^2,...,\bar{L}_j^h]$, yield $\bar{L}_j \in \mathbb{R}^{B \times T, H \times W, F}$.

The sequential GloFormer and LoFormer blocks are repeated four times to capture both high and low frequency features. To determine the SIC output after the final repetition, $\hat{Y}_j \in \mathbb{R}^{B, 1,H0,W0}$, the LoFormer attention matrix $\bar{L}_j$ is reshaped to $Z_j \in \mathbb{R}^{4, B, T, F \times H \times W}$ and an interpolation head is applied.

The primary training strategy is a geographically-weighted L1 loss that compares the predicted SIC, $\hat{Y_i}$, to the NT SIC, $Y_i$, for $i \in \{1, ..., N\}$, where $N$ is the total number of training samples, as shown below.
\begin{equation}
\mathcal{L}_{L1-GW}= \sum_{i=1}^{N} GW * \lVert \hat{Y}_i - Y_i \rVert_{1}
\label{L1_eq}
\end{equation}
The weight of each sample ($GW$) is determined according to the U.S NIC ice charts, where more confident regions like open water and ice pack are weighted higher than the less confident marginal ice zone. The L1 loss is also used to determine the best epoch during validation.

\subsection{Bayesian Transformer Model}

The high-resolution Transformer model becomes a Bayesian Neural Network (BNN) that can quantify epistemic uncertainty by assuming a probability distribution over the model parameters, $\omega$ \cite{ABDAR2021243}.
 For this study, a Gaussian likelihood for regression,  $p(y \mid x, \omega)$, is defined as follows:

\begin{equation}
    p(y \mid x, \omega) = \mathcal{N}\big(y \,;\, f^{\omega}(x), \tau^{-1} I\big)
    \label{gaussian_likelihood_eq}
\end{equation}
Where $\tau$ is the model precision. By applying Bayes theorem, the posterior distribution for the training dataset \(\mathcal{D}=\{X,Y\}\) can be determined as follows:
\begin{equation}
    p(\omega \mid X,Y) = \frac{p(Y|X, \omega) \, p(\omega)}{p(Y|X)}
\end{equation}
The posterior distribution, $p(\omega \mid X,Y)$, is approximated by variational distribution, $q_{\theta}(\omega)$, as it cannot be derived explicitly. Therefore, the secondary learning objective of the Bayesian Transformer is to approximate a distribution that is close to the true posterior distribution found by the model through BBB, with respect to the variational parameters, $\theta$. Assuming a diagonal Gaussian variational posterior, $p(\omega) = \mathcal{N}(0, I)$, the Kullback-Leibler (KL) divergence loss is used to minimize the difference between the two distributions as follows: 
\begin{equation}
\mathcal{L}_{\text{KL}}
= \mathrm{KL}\!\big(q_{\theta}(\omega) \,\|\, p(\omega)\big)
= \int q_{\theta}(\omega) \, \log \frac{q_{\theta}(\omega)}{p(\omega)} \, d\omega
\end{equation}
The final loss for multitask learning is calculated as the sum of the geographically-weighted L1 loss and KL divergence loss, $\mathcal{L}=\mathcal{L}_{L1-GW} + \mathcal{L}_{KL}$. After training, the approximate variational distribution is randomly sampled to obtain the model parameters for a set number of inferences. The predictive SIC sample mean, $\mu(x)$, and variance (that is used to quantify uncertainties), $\sigma^{2}(x)$, are derived from these inferences as follows:
\begin{equation}
\mu(x) := \mathbb{E}_{q_\theta(\omega)} \big[ f^{\omega}(x) \big]
\end{equation}
\begin{equation}
\sigma^{2}(x) := \operatorname{Var}_{q_\theta(\omega)} \big[ f^{\omega}(x) \big]
\label{eq_variance}
\end{equation}

\section{Results and Analysis}
\label{experiments}

\begin{figure*}[!htbp]
        \centering
        \includegraphics[scale = 0.4]{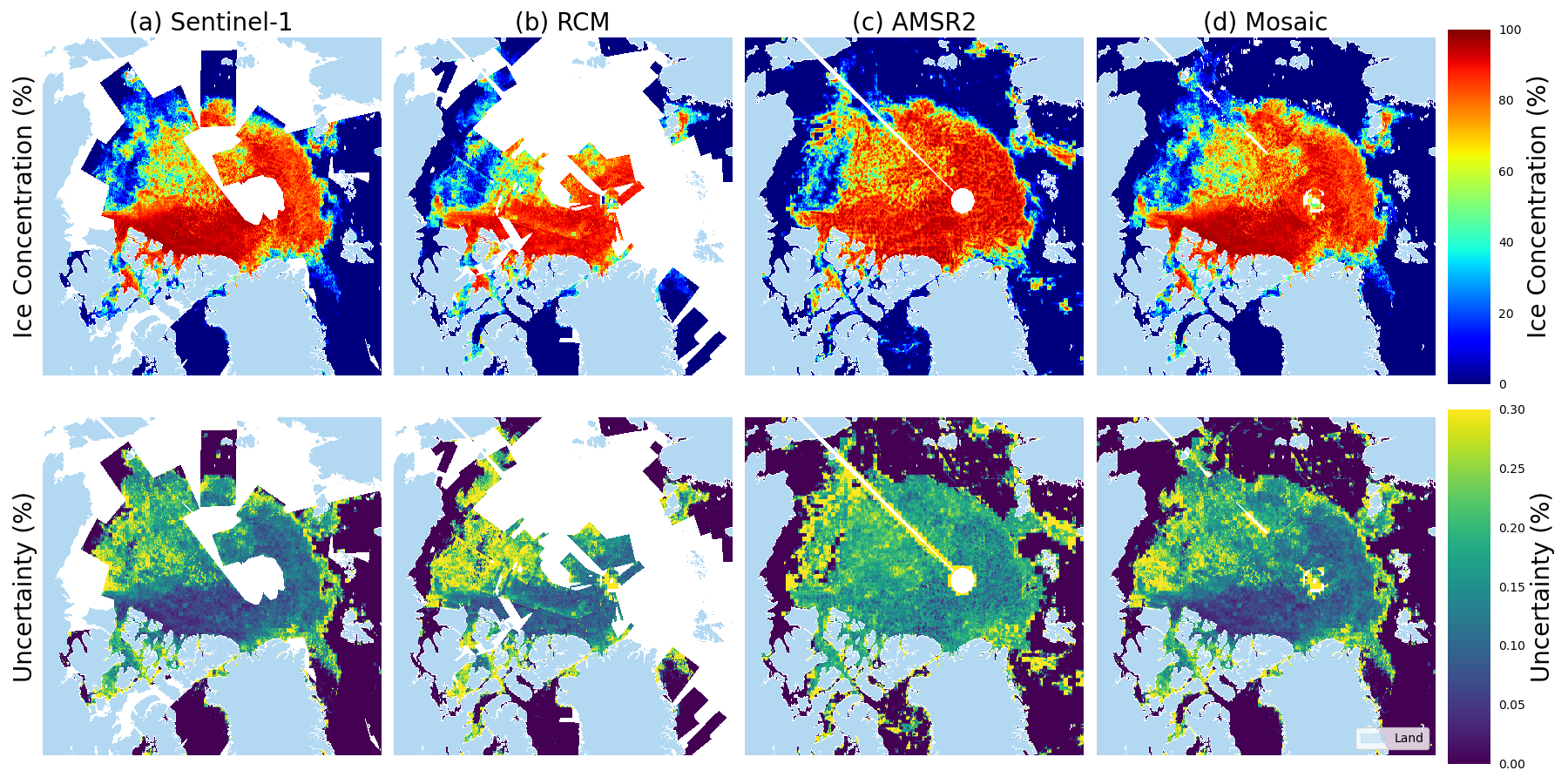}
        \vspace{-0.5cm}
        \caption{Pan-Arctic SIC (top) and corresponding uncertainty (bottom) from the High-Resolution Bayesian Transformer for September 4th, 2021. SIC and uncertainty are derived from (a) Sentinel-1 (200m resolution), (b) RCM (200m resolution), and (c) AMSR2 (5km resolution), and (d) is the final mosaic of all data sources with Sentinel-1 layered on top, followed by RCM and AMSR2.}
        \label{fig:ic_maps}
\end{figure*}

Figure \ref{fig:ic_maps} demonstrates that the proposed Bayesian Transformer model has greater uncertainties in the marginal ice zone than in open water or ice pack, which is supported by previous literature \cite{chen2023predicting}. The marginal ice zone is highly dynamic and changes rapidly, and its subtle sea ice signatures in SAR imagery can easily be mistaken for wind effects or noise over open water. Figure \ref{fig:ic_maps} also demonstrate that the proposed Bayesian Transformer model can successfully fuse different data types for pan-Arctic SIC mapping and uncertainty quantification. 

Table \ref{tab:uncert_tab} shows consistent conclusions with Figure \ref{fig:ic_maps}. The marginal ice zone (10\% to 90\%) has greater uncertainty than open water (0\% to 10\%) and ice pack (90\% to 100\%) for all methods and data sources. When comparing different datasets, the AMSR2 and RCM-derived SIC have greater uncertainty than Sentinel-1-derived SIC for all methods, likely due to the cloud and fog that impacts the 89.0 GHz AMSR2 channel and the thermal noise and incident angle effect more common in RCM. Overall, the proposed Bayesian Transformer produced the lowest model uncertainty with Sentinel-1, as shown in the highlighted Table \ref{tab:uncert_tab} row. Therefore, our choice to place the AMSR2 and RCM SIC below the Sentinel-1 SIC during fusion is supported by the uncertainties given by the Bayesian Transformer model. 

When comparing uncertainty quantification methods, Table \ref{tab:uncert_tab} demonstrates how the proposed Bayesian Transformer provides smaller and more consistent model uncertainties (0.01\% - 0.41\%) across data sources than MC dropout (0.17\% - 29.16\%) and epoch ensemble (0.64\% - 14.29\%). As the Bayesian Transformer integrates uncertainty estimation into the training objective, it can produce uncertainty estimates that are more consistent across heterogeneous data than MC dropout and epoch ensemble. In addition, the Bayesian Transformer can mitigate variance as its Bayesian prior acts as a regularizer. The Bayesian Transformer thus offers more reliable uncertainty quantification for SIC mapping using heterogeneous datasets compared to other common approaches.

\begin{table*}[t]
\renewcommand{\arraystretch}{1.2}
\centering
\caption{Mean Uncertainty (\%) per Ice Concentration Class Achieved by Different Methods on the Test Data}
\begin{tabular}{c|c|llllllllll}
                                           &                                & \multicolumn{10}{c}{Ice   Concentration Classes (\%)}                                                                                                                                                                                                                                                                                                                                                                                                                                                            \\ \cline{3-12} 
\multirow{-2}{*}{Uncertainty   Estimation} & \multirow{-2}{*}{Input   Data} & \multicolumn{1}{c|}{0-10}                         & \multicolumn{1}{c|}{10-20}                        & \multicolumn{1}{c|}{20-30}                        & \multicolumn{1}{c|}{30-40}                        & \multicolumn{1}{c|}{40-50}                        & \multicolumn{1}{c|}{50-60}                        & \multicolumn{1}{c|}{60-70}                        & \multicolumn{1}{c|}{70-80}                        & \multicolumn{1}{c|}{80-90}                        & \multicolumn{1}{c}{90-100}   \\ \hline
                                           & Sentinel-1                     & \multicolumn{1}{l|}{0.17}                         & \multicolumn{1}{l|}{5.38}                         & \multicolumn{1}{l|}{5.59}                         & \multicolumn{1}{l|}{5.35}                         & \multicolumn{1}{l|}{4.85}                         & \multicolumn{1}{l|}{4.37}                         & \multicolumn{1}{l|}{3.68}                         & \multicolumn{1}{l|}{2.65}                         & \multicolumn{1}{l|}{1.44}                         & 0.82                         \\
                                           & RCM                            & \multicolumn{1}{l|}{0.18}                         & \multicolumn{1}{l|}{7.53}                         & \multicolumn{1}{l|}{7.94}                         & \multicolumn{1}{l|}{8.94}                         & \multicolumn{1}{l|}{8.86}                         & \multicolumn{1}{l|}{7.80}                         & \multicolumn{1}{l|}{6.15}                         & \multicolumn{1}{l|}{4.13}                         & \multicolumn{1}{l|}{2.07}                         & 1.05                         \\
\multirow{-3}{*}{MC Dropout}    & AMSR2                          & \multicolumn{1}{l|}{7.91}                         & \multicolumn{1}{l|}{18.30}                        & \multicolumn{1}{l|}{22.57}                        & \multicolumn{1}{l|}{25.17}                        & \multicolumn{1}{l|}{27.78}                        & \multicolumn{1}{l|}{29.16}                        & \multicolumn{1}{l|}{26.85}                        & \multicolumn{1}{l|}{21.35}                        & \multicolumn{1}{l|}{13.68}                        & 4.93                         \\ \hline
                                           & Sentinel-1                     & \multicolumn{1}{l|}{0.72}                         & \multicolumn{1}{l|}{11.69}                        & \multicolumn{1}{l|}{12.65}                        & \multicolumn{1}{l|}{12.96}                        & \multicolumn{1}{l|}{12.52}                        & \multicolumn{1}{l|}{11.37}                        & \multicolumn{1}{l|}{9.49}                         & \multicolumn{1}{l|}{6.85}                         & \multicolumn{1}{l|}{3.80}                         & 1.94                         \\
                                           & RCM                            & \multicolumn{1}{l|}{0.64}                         & \multicolumn{1}{l|}{10.64}                        & \multicolumn{1}{l|}{12.31}                        & \multicolumn{1}{l|}{14.03}                        & \multicolumn{1}{l|}{14.29}                        & \multicolumn{1}{l|}{13.32}                        & \multicolumn{1}{l|}{11.45}                        & \multicolumn{1}{l|}{8.54}                         & \multicolumn{1}{l|}{4.45}                         & 2.45                         \\
\multirow{-3}{*}{Epoch   Ensemble}         & AMSR2                          & \multicolumn{1}{l|}{0.88}                         & \multicolumn{1}{l|}{11.39}                        & \multicolumn{1}{l|}{12.72}                        & \multicolumn{1}{l|}{13.93}                        & \multicolumn{1}{l|}{13.71}                        & \multicolumn{1}{l|}{12.21}                        & \multicolumn{1}{l|}{9.86}                         & \multicolumn{1}{l|}{6.54}                         & \multicolumn{1}{l|}{3.50}                         & 2.16                         \\ \hline
                                           & Sentinel-1                     & \multicolumn{1}{l|}{\cellcolor[HTML]{DBE5F1}0.01} & \multicolumn{1}{l|}{\cellcolor[HTML]{DBE5F1}0.30} & \multicolumn{1}{l|}{\cellcolor[HTML]{DBE5F1}0.32} & \multicolumn{1}{l|}{\cellcolor[HTML]{DBE5F1}0.33} & \multicolumn{1}{l|}{\cellcolor[HTML]{DBE5F1}0.32} & \multicolumn{1}{l|}{\cellcolor[HTML]{DBE5F1}0.29} & \multicolumn{1}{l|}{\cellcolor[HTML]{DBE5F1}0.24} & \multicolumn{1}{l|}{\cellcolor[HTML]{DBE5F1}0.19} & \multicolumn{1}{l|}{\cellcolor[HTML]{DBE5F1}0.13} & \cellcolor[HTML]{DBE5F1}0.07 \\
                                           & RCM                            & \multicolumn{1}{l|}{0.01}                         & \multicolumn{1}{l|}{0.29}                         & \multicolumn{1}{l|}{0.34}                         & \multicolumn{1}{l|}{0.35}                         & \multicolumn{1}{l|}{0.34}                         & \multicolumn{1}{l|}{0.31}                         & \multicolumn{1}{l|}{0.26}                         & \multicolumn{1}{l|}{0.20}                         & \multicolumn{1}{l|}{0.13}                         & 0.08                         \\
\multirow{-3}{*}{Bayesian   Transformer}   & AMSR2                          & \multicolumn{1}{l|}{0.02}                         & \multicolumn{1}{l|}{0.41}                         & \multicolumn{1}{l|}{0.39}                         & \multicolumn{1}{l|}{0.39}                         & \multicolumn{1}{l|}{0.37}                         & \multicolumn{1}{l|}{0.33}                         & \multicolumn{1}{l|}{0.28}                         & \multicolumn{1}{l|}{0.23}                         & \multicolumn{1}{l|}{0.19}                         & 0.13                        
\end{tabular}
\label{tab:uncert_tab}
\end{table*}

\begin{figure}[!t]
    \centering
    \includegraphics[width=\linewidth]{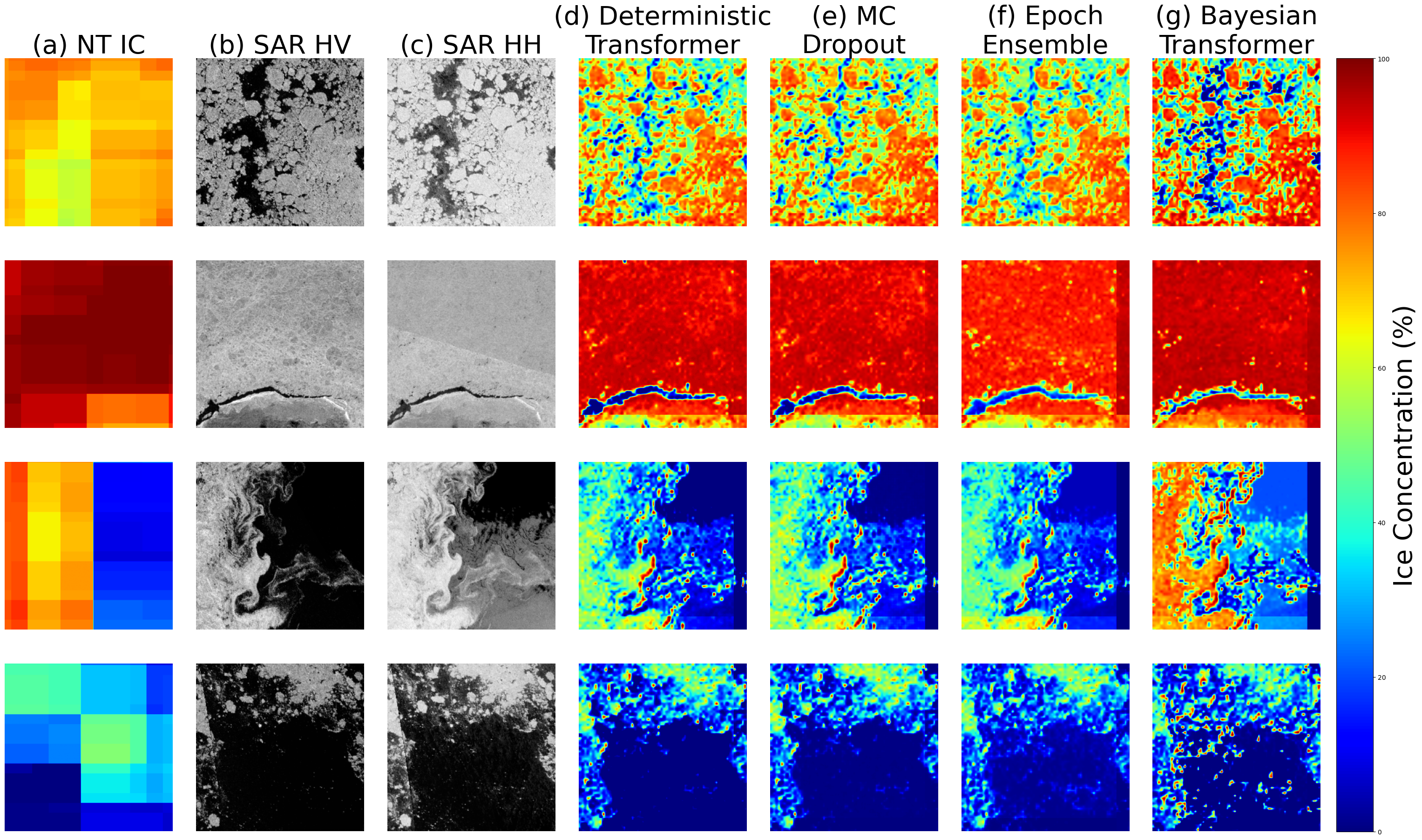}
    \vspace{-0.5cm}
    \caption{Local visual comparison of SIC derived from Sentinel-1 on September 4th, 2021, where (a) NASA Team SIC, (b) Sentinel-1 HV Imagery, (c) Sentinel-1 HH Imagery, (d) Deterministic Transformer SIC, (e) Mean Monte Carlo Dropout SIC, (f) Mean Epoch Ensemble SIC, and \textbf{(g) Mean Bayesian Transformer SIC (our approach)}.}
    \label{fig:S1_compare}
\end{figure}

\begin{figure}[!t]
    \centering
    \includegraphics[width=\linewidth]{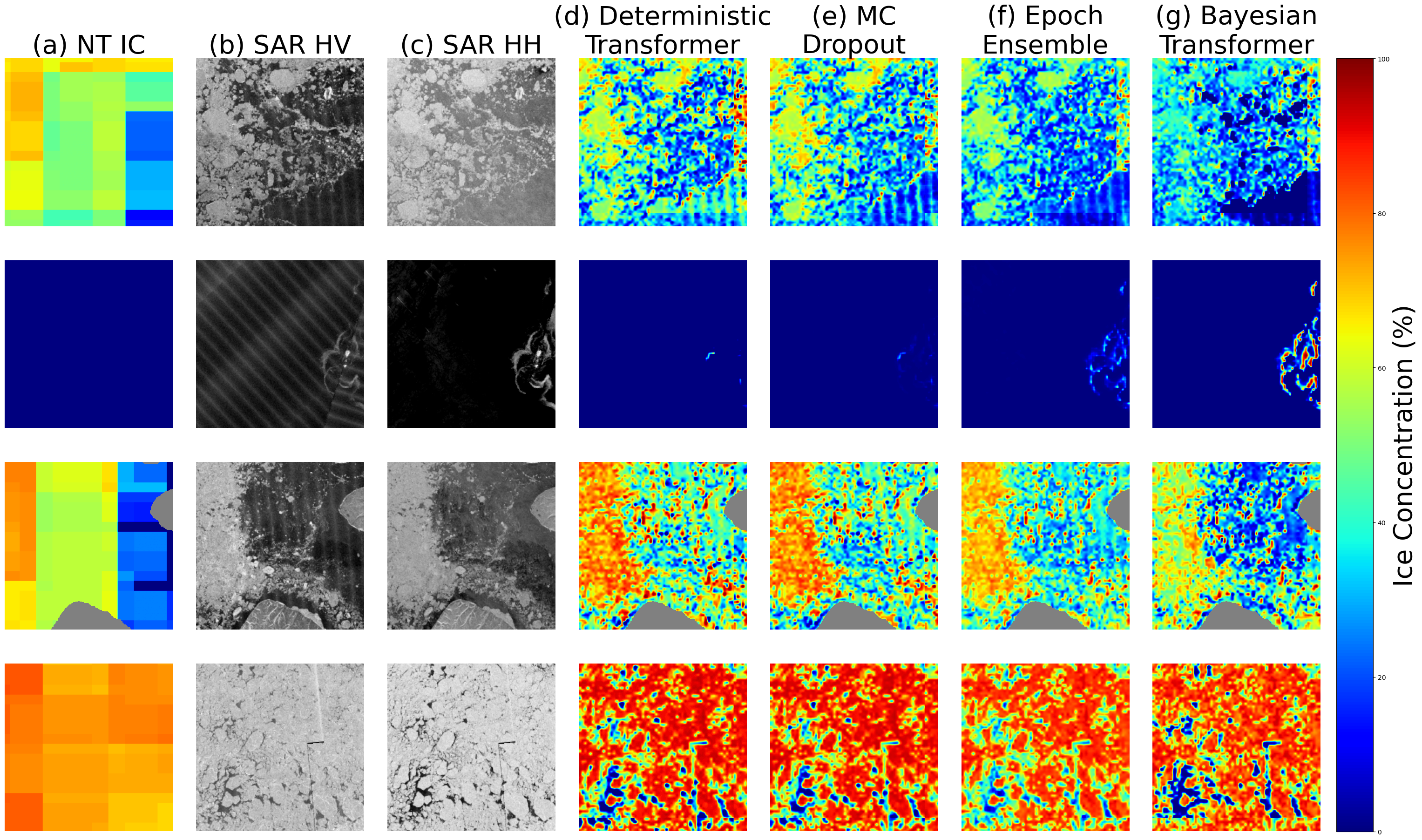}
    \vspace{-0.5cm}
    \caption{Local visual comparison of SIC derived from RCM on September 4th, 2021, where (a) NASA Team SIC, (b) RCM HV Imagery, (c) RCM HH Imagery, (d) Deterministic Transformer SIC, (e) Mean Monte Carlo Dropout SIC, (f) Mean Epoch Ensemble SIC, and \textbf{(g) Mean Bayesian Transformer SIC (our approach)}.}
    \label{fig:RCM_compare}
\end{figure}

Figures \ref{fig:S1_compare} and \ref{fig:RCM_compare} demonstrate that the stochastic model parameters in the proposed Bayesian Transformer can also improve SIC accuracy. The proposed Bayesian Transformer is the most consistent with the NT SIC (Figure \ref{fig:S1_compare} row 3 and Figure \ref{fig:RCM_compare} row 3) while also identifying the most small ice features in the SAR imagery (Figure \ref{fig:S1_compare} row 4 and Figure \ref{fig:RCM_compare} row 2). Additionally, the proposed Bayesian Transformer mitigates the impact of wind effects (Figure \ref{fig:S1_compare} row 3) and thermal noise (Figure \ref{fig:RCM_compare} row 1, row 2, and row 3) on SIC estimates. This shows how regularization through priors and explicit uncertainty estimation during learning with the Bayesian Transformer can effectively reduce overfitting to noisy or erroneous training samples for SIC mapping. Overall, all methods that utilize the Transformer model produce higher resolution SIC than the original NT SIC product and preserve many of the small ice features that are present in the SAR imagery.

Table \ref{tab:acc_tab} shows that the proposed Bayesian Transformer achieved the highest accuracy for Sentinel-1 and AMSR2, whereas MC dropout outperformed the other uncertainty quantification methods for RCM. The Bayesian Transformer achieves higher SIC accuracy than the deterministic Transformer, MC dropout, and epoch ensemble for majority of the datasets.

\begin{table}[t]
\renewcommand{\arraystretch}{1.2}
\centering
\caption{Sea Ice Concentration Accuracy Calculated using 100 high quality image chips from each validation date.}
\begin{tabular}{c|c|c|c|c}
\begin{tabular}[c]{@{}c@{}}Input\\ Data\end{tabular} & Method                                                              & $R^2$    & RMSE (\%) & MAE (\%) \\ \hline
\multirow{4}{*}{Sentinel-1}                          & \begin{tabular}[c]{@{}c@{}}Deterministic\\ Transformer\end{tabular} & 0.846 & 11.588   & 8.793   \\ \cline{2-5} 
                                                     & \begin{tabular}[c]{@{}c@{}}MC Dropout\end{tabular}         & 0.861 & 11.006   & 8.287   \\ \cline{2-5} 
                                                     & \begin{tabular}[c]{@{}c@{}}Epoch Ensemble\end{tabular}       & 0.864 & 10.905   & 8.258   \\ \cline{2-5} 
                                                     & \begin{tabular}[c]{@{}c@{}}\textbf{Bayesian}\\ \textbf{Transformer}\end{tabular} & \textbf{0.884} & \textbf{10.060}   & \textbf{7.711}   \\ \hline
\multirow{4}{*}{RCM}                                 & \begin{tabular}[c]{@{}c@{}}Deterministic\\ Transformer\end{tabular} & 0.782 & 12.591   & 9.738   \\ \cline{2-5} 
                                                     & \begin{tabular}[c]{@{}c@{}}\textbf{MC Dropout}\end{tabular}         & \textbf{0.782} & \textbf{12.568}   & \textbf{9.700}   \\ \cline{2-5} 
                                                     & \begin{tabular}[c]{@{}c@{}}Epoch Ensemble\end{tabular}       & 0.514 & 18.775   & 13.366  \\ \cline{2-5} 
                                                     & \begin{tabular}[c]{@{}c@{}}Bayesian\\ Transformer\end{tabular} & 0.750 & 13.481   & 10.765  \\ \hline
\multirow{4}{*}{AMSR2}                               & \begin{tabular}[c]{@{}c@{}}Deterministic\\ Transformer\end{tabular} & 0.865 & 10.848   & 7.555   \\ \cline{2-5} 
                                                     & \begin{tabular}[c]{@{}c@{}}MC Dropout\end{tabular}         & 0.223 & 26.039   & 21.478  \\ \cline{2-5} 
                                                     & \begin{tabular}[c]{@{}c@{}}Epoch Ensemble\end{tabular}       & 0.875 & 10.424   & 7.398   \\ \cline{2-5} 
                                                     & \begin{tabular}[c]{@{}c@{}}\textbf{Bayesian}\\ \textbf{Transformer}\end{tabular} & \textbf{0.902} & \textbf{9.264}    & \textbf{7.043}  
\end{tabular}
\label{tab:acc_tab}
\end{table}

\section{Conclusion}
\label{conclusion}
Accurate daily Pan-Arctic SIC mapping with corresponding uncertainty quantification is crucial for monitoring climate change, supporting Northern Indigenous communities, and ensuring safe navigation as sea ice extents continue to decrease. However, this is a challenging task due to the subtle nature of ice signature features, model ambiguity, and data heterogeneity. In this letter, we proposed a high-resolution Bayesian Transformer with global (GloFormer) and local (LoFormer) modules for improved feature extraction and a built-in Bayesian framework to better capture uncertainty compared to alternative approaches. We performed decision-level fusion of Sentinel-1, RCM, and AMSR2 data to address data heterogeneity and enhance Pan-Arctic SIC mapping. Overall, the Bayesian Transformer can effectively mitigate variance across heterogenous datasets and has stronger generalization capabilities than the deterministic Transformer, allowing for more accurate detection of small ice leads and floes. Future work for this study will explore other advanced DL models for SIC mapping, model and data uncertainty quantification using Heteroscedastic Bayesian Neural Networks (HBNN) \cite{chen2023predicting,chen2023uncertainty}, and multi-modal frameworks for feature-level data fusion of SAR and PM systems. Understanding the inherent uncertainty in DL-based SIC estimation will support long-term, continuous mapping of Pan-Arctic sea ice, providing a reliable baseline for future environmental studies of the Arctic and cryosphere.

\section{Acknowledgement}
The authors acknowledge the Government of Canada Earth Observation Data Management System (EODMS), Japan Aerospace Exploration Agency (JAXA), European Space Agency (ESA), and the U.S. National Ice Center for providing the datasets in this study. This work was supported in part by the Natural Sciences and Engineering Research Council (NSERC) of Canada.

\bibliographystyle{IEEEtran}

\bibliography{bibtex/bib/IEEEabrv,bibtex/uncertainty_ref.bib}

\begin{thebibliography}{10}
\providecommand{\url}[1]{#1}
\csname url@samestyle\endcsname
\providecommand{\newblock}{\relax}
\providecommand{\bibinfo}[2]{#2}
\providecommand{\BIBentrySTDinterwordspacing}{\spaceskip=0pt\relax}
\providecommand{\BIBentryALTinterwordstretchfactor}{4}
\providecommand{\BIBentryALTinterwordspacing}{\spaceskip=\fontdimen2\font plus
\BIBentryALTinterwordstretchfactor\fontdimen3\font minus \fontdimen4\font\relax}
\providecommand{\BIBforeignlanguage}[2]{{%
\expandafter\ifx\csname l@#1\endcsname\relax
\typeout{** WARNING: IEEEtran.bst: No hyphenation pattern has been}%
\typeout{** loaded for the language `#1'. Using the pattern for}%
\typeout{** the default language instead.}%
\else
\language=\csname l@#1\endcsname
\fi
#2}}
\providecommand{\BIBdecl}{\relax}
\BIBdecl

\bibitem{wang2016sea}
L.~Wang, K.~A. Scott, L.~Xu, and D.~A. Clausi, ``Sea ice concentration estimation during melt from dual-pol sar scenes using deep convolutional neural networks: A case study,'' \emph{IEEE Transactions on Geoscience and Remote Sensing}, vol.~54, no.~8, pp. 4524--4533, 2016.

\bibitem{malmgren2021convolutional}
D.~Malmgren-Hansen, L.~T. Pedersen, A.~A. Nielsen, M.~B. Kreiner, R.~Saldo, H.~Skriver, J.~Lavelle, J.~Buus-Hinkler, and K.~H. Krane, ``A convolutional neural network architecture for sentinel-1 and amsr2 data fusion,'' \emph{IEEE Transactions on Geoscience and Remote Sensing}, vol.~59, no.~3, pp. 1890--1902, 2021.

\bibitem{wulf2024panarctic}
\BIBentryALTinterwordspacing
T.~Wulf, J.~Buus-Hinkler, S.~Singha, H.~Shi, and M.~B. Kreiner, ``Pan-arctic sea ice concentration from sar and passive microwave,'' \emph{The Cryosphere}, vol.~18, no.~11, pp. 5277--5300, 2024. [Online]. Available: \url{https://tc.copernicus.org/articles/18/5277/2024/}
\BIBentrySTDinterwordspacing

\bibitem{He2025physically}
J.~He, Y.~Zhao, D.~Yang, H.~Wang, and X.~Deng, ``Physically constrained spatiotemporal deep learning model for fine-scale, long-term arctic sea ice concentration prediction,'' \emph{IEEE Transactions on Geoscience and Remote Sensing}, vol.~63, pp. 1--21, 2025.

\bibitem{ABDAR2021243}
\BIBentryALTinterwordspacing
M.~Abdar, F.~Pourpanah, S.~Hussain, D.~Rezazadegan, L.~Liu, M.~Ghavamzadeh, P.~Fieguth, X.~Cao, A.~Khosravi, U.~R. Acharya, V.~Makarenkov, and S.~Nahavandi, ``A review of uncertainty quantification in deep learning: Techniques, applications and challenges,'' \emph{Information Fusion}, vol.~76, pp. 243--297, 2021. [Online]. Available: \url{https://www.sciencedirect.com/science/article/pii/S1566253521001081}
\BIBentrySTDinterwordspacing

\bibitem{aires2004neural}
F.~Aires, C.~Prigent, and W.~B. Rossow, ``Neural network uncertainty assessment using bayesian statistics: A remote sensing application,'' \emph{Neural computation}, vol.~16, no.~11, pp. 2415--2458, 2004.

\bibitem{asadi2020evaluation}
N.~Asadi, K.~A. Scott, A.~S. Komarov, M.~Buehner, and D.~A. Clausi, ``Evaluation of a neural network with uncertainty for detection of ice and water in sar imagery,'' \emph{IEEE Transactions on Geoscience and Remote Sensing}, vol.~59, no.~1, pp. 247--259, 2020.

\bibitem{wang2016improved}
\BIBentryALTinterwordspacing
L.~Wang, K.~A. Scott, and D.~A. Clausi, ``Improved sea ice concentration estimation through fusing classified sar imagery and amsr-e data,'' \emph{Canadian Journal of Remote Sensing}, vol.~42, no.~1, pp. 41--52, 2016. [Online]. Available: \url{https://doi.org/10.1080/07038992.2016.1152547}
\BIBentrySTDinterwordspacing

\bibitem{NTdetails}
\BIBentryALTinterwordspacing
D.~J. Cavalieri, P.~Gloersen, and W.~J. Campbell, ``Determination of sea ice parameters with the nimbus 7 smmr,'' \emph{Journal of Geophysical Research: Atmospheres}, vol.~89, no.~D4, pp. 5355--5369, 1984. [Online]. Available: \url{https://agupubs.onlinelibrary.wiley.com/doi/abs/10.1029/JD089iD04p05355}
\BIBentrySTDinterwordspacing

\bibitem{chen2023predicting}
X.~Chen, R.~Valencia, A.~Soleymani, and K.~A. Scott, ``Predicting sea ice concentration with uncertainty quantification using passive microwave and reanalysis data: A case study in baffin bay,'' \emph{IEEE Transactions on Geoscience and Remote Sensing}, vol.~61, pp. 1--13, 2023.

\bibitem{chen2023uncertainty}
X.~Chen, K.~A. Scott, L.~Xu, M.~Jiang, Y.~Fang, and D.~A. Clausi, ``Uncertainty-incorporated ice and open water detection on dual-polarized sar sea ice imagery,'' \emph{IEEE Transactions on Geoscience and Remote Sensing}, vol.~61, pp. 1--13, 2023.

\end{thebibliography}

\end{document}